\documentclass[10pt,twocolumn,letterpaper]{article}

\usepackage[pagenumbers]{cvpr} 
\usepackage{multirow}
\usepackage{adjustbox}
\usepackage{pifont}
\usepackage{amsfonts}

\usepackage{microtype}

\definecolor{cvprblue}{rgb}{0.21,0.49,0.74}
\usepackage[pagebackref,breaklinks,colorlinks,allcolors=cvprblue]{hyperref}

\def\paperID{N/A} 
\def\confName{CVPR}
\def\confYear{2026}

\title{Incentivizing Cardiologist-Like Reasoning in MLLMs for Interpretable Echocardiographic Diagnosis}

\author{Yi Qin$^1$, Lehan Wang$^1$, Chenxu Zhao$^2$, Alex P.W. Lee$^2$, Xiaomeng Li$^1$\thanks{Corresponding to Xiaomeng Li (eexmli@ust.hk).} \\ $^1$The Hong Kong University of Science and Technology \\ $^2$ The Chinese University of Hong Kong
}

\begin{document}
\maketitle
\newcommand{\methodname}{CardiacMind}
\newcommand{\reasonname}{ECT-RL}
\newcommand{\token}[1]{{\footnotesize\texttt{#1}}}
\newcommand{\lh}[1]{{\color[rgb]{0.5,0.3,0.8}{[LH: #1]}}}
\newcommand{\xmli}[1]{{\color[rgb]{0.9,0.0,0.0}{[XM: #1]}}}
\newcommand{\qin}[1]{{\color[rgb]{0,0,0.8}{[QIN: #1]}}}

\newcommand{\yes}{\ding{51}}
\newcommand{\no}{\ding{55}}

\begin{abstract}
Echocardiographic diagnosis is vital for cardiac screening yet remains challenging.
Existing echocardiography foundation models do not effectively capture the relationships between quantitative measurements and clinical manifestations, whereas medical reasoning multimodal large language models (MLLMs) require costly construction of detailed reasoning paths and remain ineffective at directly incorporating such echocardiographic priors into their reasoning.
To address these limitations, we propose a novel approach comprising Cardiac Reasoning Template (CRT) and \methodname{} to enhance MLLM's echocardiographic reasoning by introducing cardiologist-like mindset.
Specifically, CRT provides stepwise canonical diagnostic procedures for complex cardiac diseases to streamline reasoning path construction without the need for costly case-by-case verification.
To incentivize reasoning MLLM under CRT, we develop \methodname{}, a new reinforcement learning scheme with three novel rewards: Procedural Quantity Reward (PQtR), Procedural Quality Reward (PQlR), and Echocardiographic Semantic Reward (ESR). PQtR promotes detailed reasoning; PQlR promotes integration of evidence across views and modalities, while ESR grounds stepwise descriptions in visual content.
Our methods show a 48\% improvement in multiview echocardiographic diagnosis for 15 complex cardiac diseases and a 5\% improvement on CardiacNet-PAH over prior methods. The user study on our method's reasoning outputs shows 93.33\% clinician agreement with cardiologist-like reasoning logic.
Our code will be available.
\end{abstract}

\vspace{-2em}
\section{Introduction}
\label{sec:intro}

\begin{figure*}[htb]
	\centering
	\includegraphics[width=1.0\textwidth]{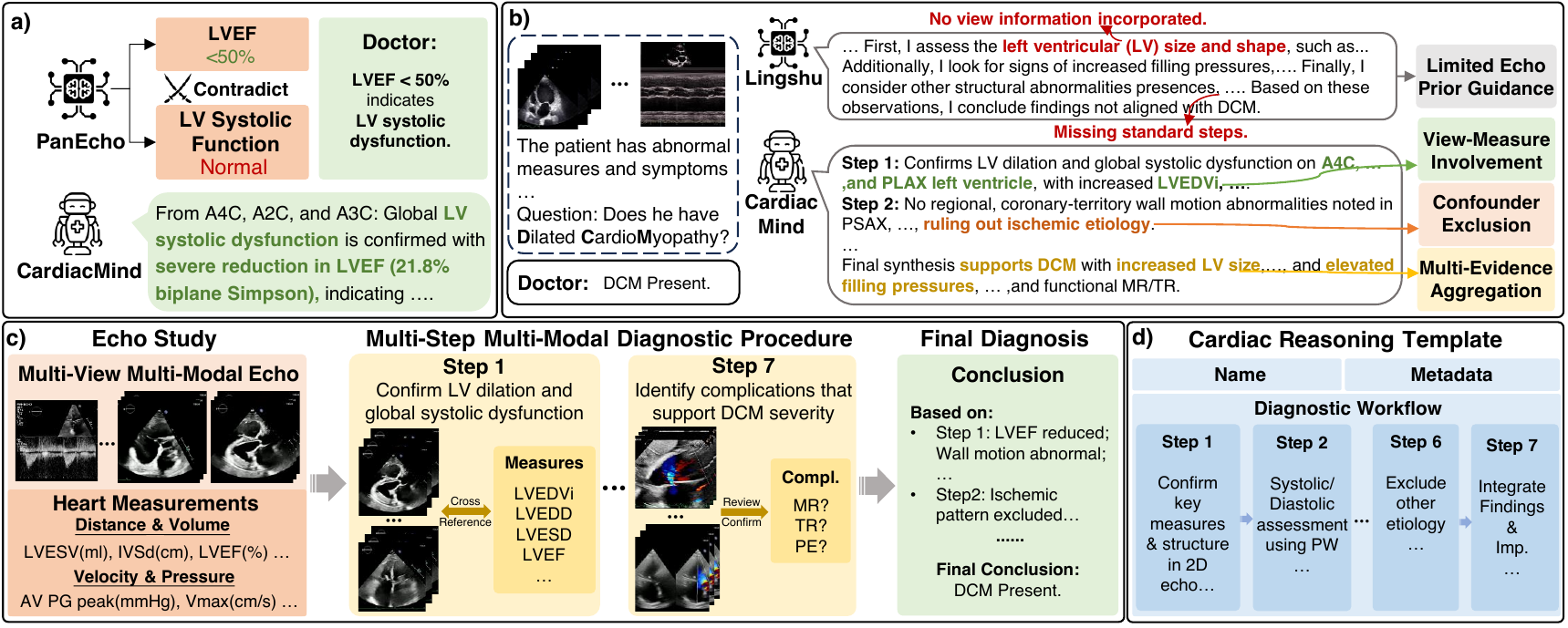}
	\caption{Comparison of \methodname{} with prior methods and overview of the echocardiography diagnosis workflow and the Cardiac Reasoning Template. a) PanEcho~\cite{holste2025complete} predictions versus our model's reasoning process. b) Reasoning process comparison between the latest medical reasoning method~\cite{xu2025lingshu} and our model. c) Summary of the dilated cardiomyopathy (DCM) clinical diagnosis workflow~\cite{bozkurt2016current}. d) Overview of the proposed Cardiac Reasoning Template.}
	\label{fig:intro}
\vspace{-1em}
\end{figure*}

Echocardiography is an essential cardiac imaging modality~\cite{wei2023variation} for assessing a wide range of cardiac diseases~\cite{ommen20242024,arbelo20232023,heidenreich20222022,stout20192018}.
Accurate diagnostic interpretation of echocardiography is challenging and requires comprehensive analysis of heterogeneous inputs~\cite{wiegers20192019}. 
This difficulty is exacerbated in complex diseases, such as septal defect subtyping~\cite{silvestry2015guidelines} or cardiomyopathy~\cite{ommen20242024}.
As~\cref{fig:intro}.c illustrates, clinicians follow a stepwise, logically ordered diagnostic procedure. This process synthesizes multimodal information (e.g., cardiac motion, chamber structure, blood jet velocities, etc.) from multiview echocardiography and quantitative measures, grounded in established cardiac knowledge.
Therefore, achieving accurate and reliable echocardiographic diagnosis remains a longstanding challenge for both clinicians and deep learning models.

Recent studies have sought to address this challenge by developing echocardiographic foundation models~\cite{holste2025complete,amadou2024echoapex,vukadinovic2025comprehensive,christensen2024vision}. However, these models still struggle with complex cardiac diseases (see~\cref{tab:main} row ``PanEcho~\cite{holste2025complete}'' and ``EchoPrime~\cite{vukadinovic2025comprehensive}'').
For instance, as shown in~\cref{fig:intro}.a, PanEcho~\cite{holste2025complete} produces contradictory outputs: quantitative measurements and disease classifications suggest opposing diagnoses, e.g., the Left Ventricular Ejection Fraction (LVEF) is below 50\%, yet the predicted Left Ventricle (LV) systolic function is labeled as normal. This inconsistency is unreasonable, as most disease classifications are strongly dependent on quantitative measurements~\cite{mitchell2019guidelines,lang2015recommendations} (e.g., an LVEF below 50\% typically indicates LV systolic dysfunction), which ultimately undermines the model’s reliability.

The main reason for this low reliability is that these models fail to capture the relationships between measurements (e.g., LVEF) and clinical manifestations (e.g., LV systolic dysfunction) when forming diagnostic conclusions. 
Medical large language models (MLLMs) have recently demonstrated promising reasoning abilities in analyzing clinical questions~\cite{wang2025proactive,huang2025elicit,su2025gmai,sun2025enhancing,pan2025medvlm,lai2025med,rui2025improving}, serving as a potential solution for integrating clinical measurements into the diagnostic decision-making process. 
However, most of these approaches were primarily developed for general medical image interpretation tasks and lack consideration of domain-specific echocardiographic knowledge, resulting in limited performance when applied directly; see results of ``MedVLM-R1~\cite{pan2025medvlm}'' and ``Chiron-o1~\cite{sun2025enhancing}'' in~\cref{tab:main}.
A straightforward solution is to incorporate these diagnostic priors by constructing and verifying case-by-case echocardiographic reasoning paths. Nevertheless, this process demands extensive effort from echocardiographers, e.g., more than 450 hours of expert time were required to build chain-of-thought data for only 100 knee scans~\cite{sambara20253dreasonknee}. Another solution is to directly provide the diagnostic priors as instructions and knowledge on diagnosis for models during inference~\cite{yang2024buffer,yang2025reasonflux}. However, we found it still achieved limited improvements in echocardiography analysis (row ``Qwen2.5-VL-ICL'' in~\cref{tab:main}).

To address these limitations, we propose a novel approach comprising a Cardiac Reasoning Template \textbf{(CRF)} and a reinforcement learning framework, namely \textbf{CardiacMind}, to better enhance MLLMs’ echocardiographic reasoning ability. Our core idea is inspired by the canonical stepwise and deductive diagnostic mindset of cardiologists (\cref{fig:intro}.c). To streamline the construction of reasoning paths, the CRF provides stepwise echocardiographic diagnostic procedures that emulate cardiologists’ diagnostic logic, incorporating knowledge distilled from authoritative sources to guide the model’s reasoning during both training and inference. 
As~\cref{fig:intro}.d shows, CRT provides accurate and concise guidance that directs the model to perform detailed analysis, addressing the previous costly data construction demand while effectively introducing echocardiographic reasoning prior.

To incentivize MLLM under CRT, we introduce \textbf{\methodname{}}, a reinforcement learning framework with three novel rewards: the Procedural Quantity Reward (\textbf{PQtR}), the Procedural Quality Reward (\textbf{PQlR}), the Echocardiography Semantic Reward (\textbf{ESR}).
\textbf{PQtR} and \textbf{PQlR} supervise the model by evaluating the extensiveness and content relevance of reasoning steps relative to the CRF, thereby promoting the logical integration of diagnostic evidence across multimodal and multiview echocardiographic inputs. 
\textbf{ESR} further strengthens the alignment between the descriptive content produced under PQlR guidance and the echocardiographic video content, enabling faithful interpretation of echocardiography.
During inference, we further introduce Template-guided Reasoning Rectification (TRR) within \methodname{}, which leverages the CRF to detect and correct deviations from procedural steps, thereby refining the model’s final diagnostic conclusions.
To summarize, our key contributions are listed as follows:
\begin{itemize}
    \item We introduce Cardiac Reasoning Template (CRT), the first high-level concise echocardiographic diagnostic reasoning path for 15 types of complex cardiac diseases, acting as the anchor for the echocardiographic reasoning model.
    \item We propose \methodname{}, a new rule-based reinforcement learning method that leverages CRT to incentivize MLLM's rigid cardiologist-like diagnostic reasoning through three novel rewards.
    \item Our results demonstrate a 48\% accuracy improvement in multiview, multimodal echocardiographic diagnosis for 15 complex cardiac diseases and a 5\% accuracy improvement on single-view CardiacNet-PAH benchmarks over prior methods. The user study on our method's reasoning outputs shows 93.33\% clinician agreement with cardiologist-like reasoning logic.
\end{itemize}

\begin{figure*}[htb]
	\centering
	\includegraphics[width=1.0\textwidth]{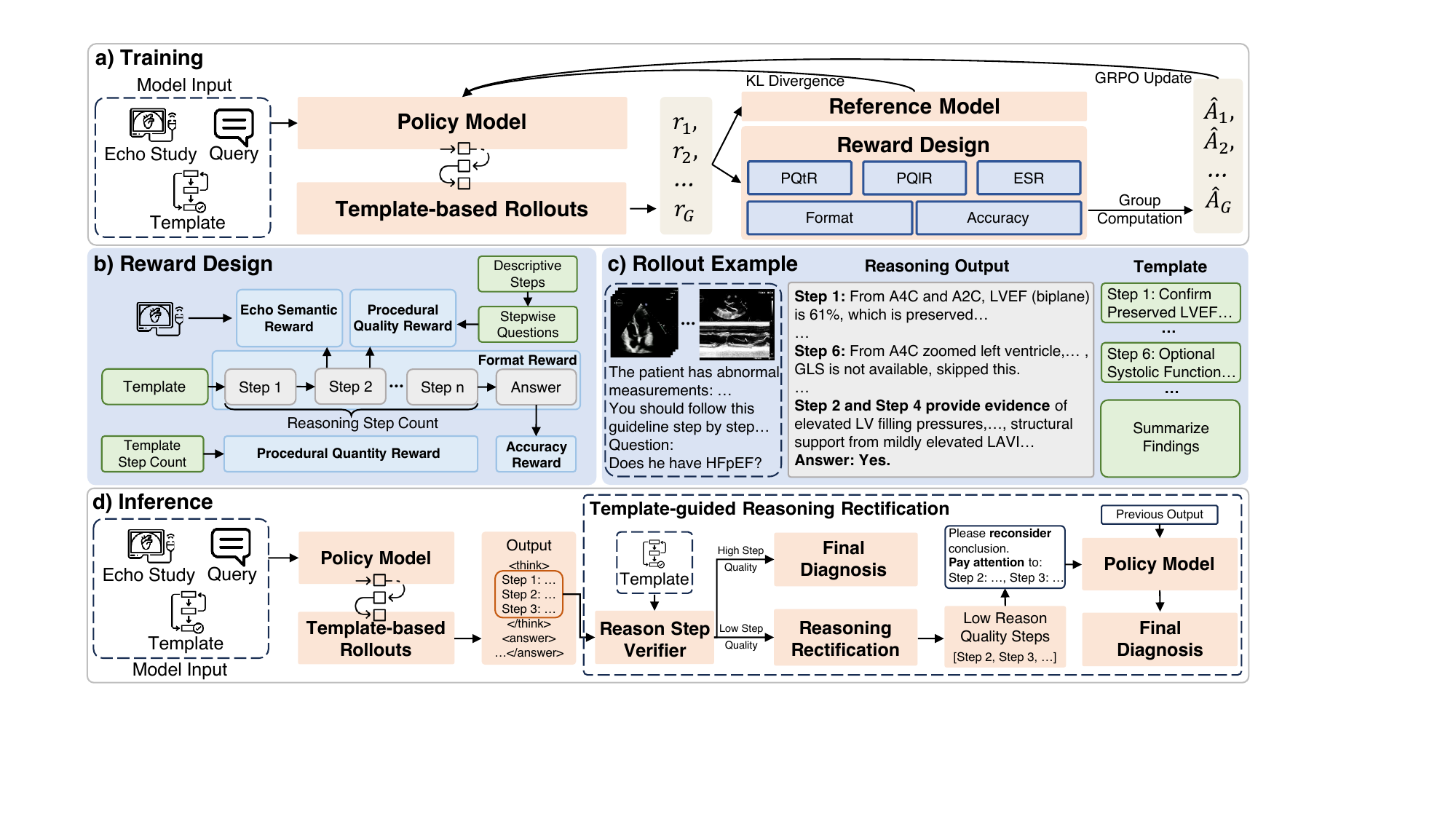}
	\caption{Overview of \methodname{}. a) Training. \methodname{} uses Group Relative Policy Optimization with three novel rewards and two basic rewards. b) Reward design. We introduce three new rewards that encourage extensive reasoning aligned with standard diagnostic procedures. Elements in Green indicate contents from the CRT template. c) Example of the reasoning process. d) Inference. \methodname{} supports scalable inference through template retrieval and Template-guided Reasoning Rectification (TRR). TRR monitors and corrects steps that deviate the prescribed procedure and it refines the final conclusion.}
	\label{fig:main}
\end{figure*}

\section{Related Works}
\label{sec:related}

\subsection{Medical Reasoning Multimodal Large Language Models}
Recent medical reasoning multimodal large language models~\cite{pan2025medvlm,su2025gmai,sun2025enhancing,huang2025elicit,lai2025med,wang2025proactive,xu2025lingshu} demonstrate improved performance on medical diagnosis by analyzing questions step by step. MedVLM-R1~\cite{pan2025medvlm}, Med-R1~\cite{lai2025med}, and GMAI-VL-R1~\cite{su2025gmai} use rule-based reward functions with Group Relative Policy Optimization (GRPO) to extend the reasoning process. MedE$^2$~\cite{huang2025elicit} constructs a reasoning database and applies Direct Preference Optimization (DPO) to further enhance reasoning. Chiron-o1~\cite{sun2025enhancing} introduces a new reasoning path search scheme to build rigorous and effective medical reasoning data. MedRwR~\cite{wang2025proactive} further improves reasoning by actively retrieving referential content during the reasoning process. 
Unlike prior work that either requires costly, case-by-case reasoning-path constructions to inject diagnostic priors~\cite{sun2025enhancing,wang2025proactive,huang2025elicit} or provides only limited supervision of the diagnostic reasoning process during training~\cite{pan2025medvlm,lai2025med,su2025gmai}, our CRT streamlines reasoning-path construction by distilling canonical stepwise diagnostic procedures into high-level, accurate guidance supporting the model’s detailed analysis.
This approach substantially reduces the cost of reasoning-path construction while effectively introducing authoritative, stepwise reasoning guidance for the model.

\subsection{Reasoning Models with Trajectory Guidance}
Recent methods guide model reasoning with external knowledge using templates~\cite{yang2025reasonflux,yang2024buffer}, process reward models~\cite{zou2025reasonflux,yun2025med}, or stepwise supervision~\cite{zhang2025r1,sambara20253dreasonknee,le2025s}. ReasonFlux retrieves hierarchical templates at inference to provide step by step guidance for math problems~\cite{yang2025reasonflux}. Med-PRM~\cite{yun2025med} uses retrieval augmented generation to verify each step against medical knowledge bases. R1-VL introduces stepwise rewards to promote accuracy and logical coherence~\cite{zhang2025r1}. 
However, directly providing such guidance only at inference for multimodal echocardiographic reasoning yields limited improvements, and constructing case-by-case guidance remains costly. In contrast, \methodname{} introduces three novel rewards within a reinforcement learning framework that encourage the model to perform detailed stepwise reasoning under the high-level instructions from CRT, without requiring case-by-case ground-truth guidance, while strengthening its association with echocardiographic content. These rewards incentivize the MLLM’s echocardiographic reasoning ability, thereby enhancing its performance under complex diseases.

\section{Methodology}
\label{sec:method}

\subsection{Cardiac Reasoning Template}
We first construct \textbf{C}ardiac \textbf{R}easoning \textbf{T}emplate (\textbf{CRT}) as the anchor reasoning trajectory for \methodname{}. 
As shown in~\cref{fig:intro}.d, each CRT template summarizes a stepwise echocardiographic diagnosis procedure distilled from textbooks and guidelines. 
Each step provides a high-level reasoning direction that guides the model to analyze and integrate echocardiographic evidence for deductive diagnosis, mirroring the workflow of cardiologists. The steps contain only high-level summaries of diagnostic procedures and exclude any patient-specific conclusions to prevent data leakage.

To construct CRT, we select 15 complex cardiac diseases to ensure the coverage of clinically relevant conditions.
We define complex cardiac diseases as conditions that require establishing a diagnosis using evidence from more than three echocardiographic views and measurements. Next, we use an automatic pipeline~\cite{QinYi_MultiAgent_MICCAI2025} to retrieve and process transthoracic echocardiography (TTE) guidelines and relevant textbooks and to organize them into a structured database. We then prompt GPT-5 to convert long, descriptive paragraphs into concise, stepwise diagnostic procedures for each disease. A board-certified cardiologist validated the CRT. We provide further details and complete examples in Supplementary~\cref{supp:CRT}.

\Cref{fig:intro}.d illustrates the CRT template structure. Each template comprises three components: Template Name, Metadata, and Diagnostic Workflow.
\textbf{Template Name} is the name of the template, denoted as $T_{name}$.
\textbf{Metadata} $T_{meta}$ contains keywords and a concise summary to facilitate retrieval. Specifically, $T_{meta}$ includes four fields: ``Knowledge Tag'', which lists key diagnostic concepts in the procedure; ``Description'', which provides a brief summary of the diagnostic process; ``Application Scenario'', which lists disease-specific abnormalities to consider; and ``Views and Measurements Required'', which lists the echocardiographic views and measurements needed for analysis. $T_{name}$ and $T_{meta}$ serve as the retrieval key for CRT.
A template is defined as $T=\{T_{name},T_{meta},T_{reason}\}$. The complete CRT is the collection of all templates, denoted as $\mathcal{T}=\{T_1,T_2,...,T_n\}$, where n is the total number of reasoning templates. We constructed $n=42$ templates in total.

During training, we retrieve the disease-specific template from CRT using $T_{name}$ and $T_{meta}$. We include the retrieved template in the model input. We design three rewards that enforce adherence to the diagnostic procedure and promote effective use of echocardiographic evidence. During inference, we use the disease query to retrieve the template with the same strategy and include it in the model input. 
CRT also enables inference scaling to improve reasoning reliability and accuracy, as detailed in~\cref{sec:inference}.

\subsection{Echocardiographic Multi-Modal Reasoning with Template Guidance}

\subsubsection{Problem Formulation}
We formulate the task as a visual question answering problem. The input is multimodal, $x=\{[v_k]_{k=1}^K, t\}$. It comprises $K$ echocardiographic inputs $[v_k]_{k=1}^K$ with view labels assigned by a strong view classification model~\cite{song2025echoviewclip}. It also includes a text component $t$ that specifies the disease query and lists echocardiographic and measurement abnormalities. This input setup mirrors clinical practice. We model the MLLM as a policy $\pi_\theta(\cdot|x,\mathrm{T})$ that performs autoregressive generation given $x$ and the retrieved reasoning template $\mathrm{T} \in \mathcal{T}$ from CRT. The policy first outputs a sequence of $i$ reasoning steps $R=\{h_1,h_2,...,h_i\}$ under template guidance. It then outputs the final disease diagnosis $y$.

{
\vspace{-1em}
\begin{figure}[htb]
	\centering
	\includegraphics[width=1.0\columnwidth]{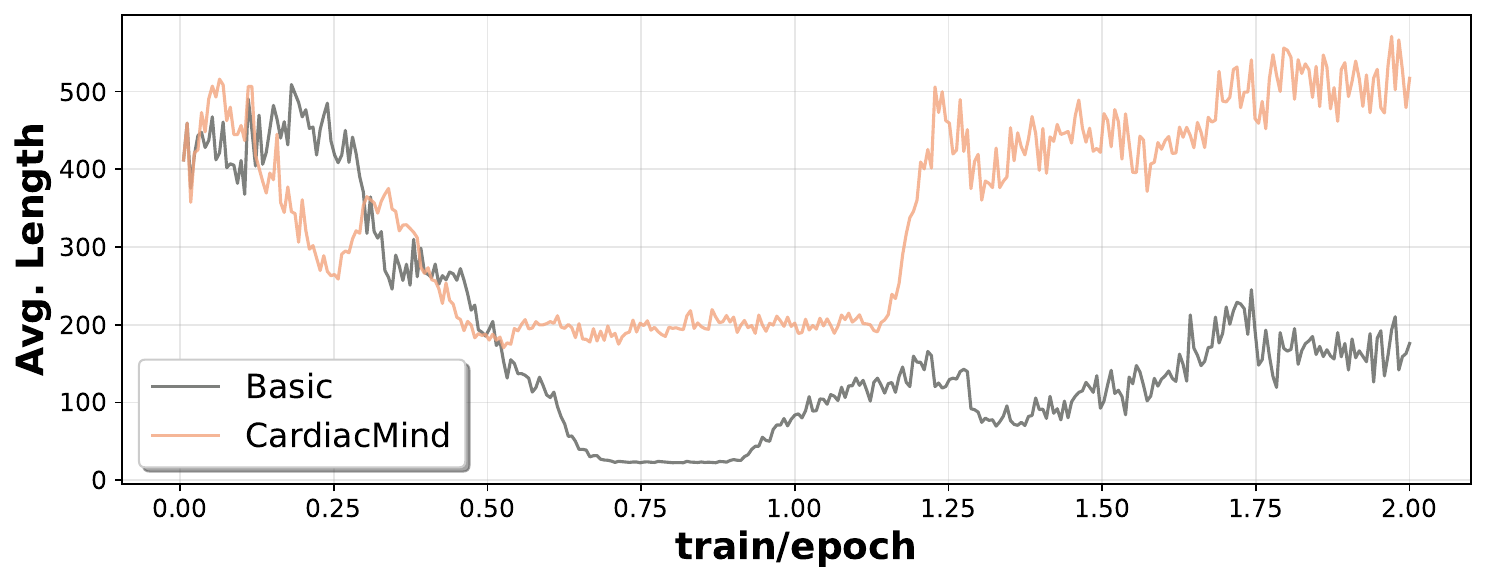}
	\caption{Average response length during training. ``Basic'' uses only accuracy and format rewards and yields short reasoning without detail on complex echocardiography inputs. \methodname{} first preserves a stepwise reasoning structure in the first training stage (epoch 0-1). It is then incentivized to produce detailed stepwise diagnostic reasoning in the second training stage (epoch 1-2).}
	\label{fig:length}
\end{figure}
\vspace{-1em}
}
\subsubsection{Reward Design}
\label{sec:reward}

We propose three rewards that leverage the curated CRT. These rewards guide the model to produce detailed and factually grounded analyses of the echocardiographic inputs with adherence to the template, as shown in~\cref{fig:main}.b.

\noindent \textbf{Procedural Quantity Reward.} We propose the Procedural Quantity Reward (PQtR) $\sigma_{PQtR}$ to encourage extensive stepwise reasoning. 
As shown in~\cref{fig:length}, complex echocardiography inputs often prevent the model from establishing an accurate and long reasoning path. In response, the model reduces its reasoning process to minimize apparent errors. Such reduction weakens logical support and can still lead to incorrect conclusions. It also undermines subsequent reward modeling for stepwise diagnostic content.
To counter this effect, PQtR counts the number of generated reasoning steps and compares it with the reference step count in the template. 
The reward encourages the model to produce explicit reasoning steps with a similar count to the reference.
We formally define PQtR in~\cref{eqn:pqtr}.

{
\vspace{-1.5em}
\begin{equation}
    \sigma_{PQtR} =
\begin{cases}
\min(1.0, \frac{|R|}{|\mathrm{T}|}), & \text{if } |R| \le |\mathrm{T}| + \epsilon \\[2pt]
0, & \text{otherwise.}
\end{cases}
\label{eqn:pqtr}
\end{equation}
\vspace{-1em}
}

In~\cref{eqn:pqtr}, $|R|$ and $|\mathrm{T}|$ denote the number of steps in the model's reasoning and in the reference template, respectively. The verbosity tolerance threshold $\epsilon$ regularizes the generation of reasoning steps and keeps the step count within a reasonable range. This allows the model to analyze and then summarize information when needed. We set $\epsilon=5$ in all experiments.

\noindent \textbf{Procedural Quality Reward.} While PQtR encourages extensive stepwise reasoning, the model must also adhere to standard diagnostic procedures and aggregate heterogeneous evidence in each step. We therefore propose the Procedural Quality Reward (PQlR) $\sigma_{PQlR}$. It measures the stepwise relevance of the generated reasoning with respect to the retrieved reasoning templates.

Specifically, for each diagnostic step $a_m$ in the retrieved templates $\mathrm{T}_{reason} \in \mathrm{T}$, we convert its description into a list of $P$ echocardiography centered questions $a_m \rightarrow [q_m^p]_{p=1}^P, P\leq5$ that target echocardiographic views and cardiac measurements (details in Supplementary~\cref{supp:qlist}). We discard questions that require views unavailable for the specific patient by using $\mathrm{T}_{meta}$. This avoids requesting nonexistent observations. Finally, for each corresponding reasoning step $h$ produced by the model $\pi_\theta$, we apply an external verifier $g(h,q) \to [0,1]$ to score $h$ for content relevance, logical coherence, and conclusiveness with respect to $q$. \Cref{eqn:pqlr} gives the formal definition of PQlR.

{
\vspace{-1.5em}
\begin{equation}
\sigma_{PQlR} =
\begin{cases}
\frac{\sum_{i=1}^{|T|} g(h_i, [q_i^p]_{p=1}^P)}{|T|}, & \text{if } |\mathrm{T}| \le |R| \le |\mathrm{T}| + \epsilon \\[2pt]
0, & \text{otherwise.}
\end{cases}
\label{eqn:pqlr}
\end{equation}
\vspace{-1.5em}
}

Here, $h_i \in R$ and $[q_i^p]_{p=1}^P \in Q$. $R$ is the set of generated reasoning steps. $Q$ is the set of question lists. The verifier $g$ produces a single score for one question set $[q_i^p]_{p=1}^P$. We compute the reward only when the total number of reasoning steps at least reaches that of the template, ensuring that the reasoning content is evaluated at the correct steps. $\epsilon$ is the verbosity tolerance threshold, identical to that used in PQtR. We implement the verifier $g$ using the LLM-as-Judge strategy~\cite{lin2023llm} with Qwen3-8B~\cite{yang2025qwen3}. See Supplementary~\cref{supp:pqlr_pmt} for details.

\noindent \textbf{Echocardiographic Semantic Reward.} To ensure faithful analysis, the echocardiographic descriptive content in the stepwise reasoning promoted by PQtR and PQlR must be aligned with the visual input.
We therefore propose the Echocardiographic Semantic Reward (ESR) $\sigma_{ESR}$ to enforce visual grounding. 
ESR encourages agreement between each video and its corresponding description in the reasoning steps by maximizing their embedding cosine similarity.
At each generated reasoning step, we identify sentences $s$ that contain echocardiographic view names that capture the descriptive content. For each such sentence, we use the view name to retrieve the corresponding video $v$ and construct the set of paired video and sentence data $V=\{(v_i,s_i)\}$. We then use a CLIP-based echocardiography visual language model~\cite{vukadinovic2025comprehensive}, denoted $f(v,s)$, as the verifier. Given a pair, $f(v,s)$ computes the text video embedding cosine similarity and normalizes it to the range $[0,1]$. The formal definition of ESR is given in Eqn.~\ref{eqn:esr}.

{
\vspace{-1em}
\begin{equation}
\sigma_{ESR} =
\begin{cases}
\frac{\sum_{i=1}^{|V|} f(v_i, s_i)}{|V|}, & \text{if } |\mathrm{T}| \le |R| \le |\mathrm{T}| + \epsilon \\[3pt]
0, & \text{otherwise.}
\end{cases}
\label{eqn:esr}
\end{equation}
}
Here, $|V|$ denotes the total number of pairs. The reward calculation threshold is identical to that used in PQlR. Note that $f(v,s)$ verifies video with single descriptive sentence instead of composite final diagnostic conclusion, thereby decreasing the difficulty of accurate text-video similarity estimation.

\noindent \textbf{Format and Accuracy Reward.} We apply basic format and accuracy rewards to enforce the output structure and the correctness of the final conclusion. For the format reward $\sigma_{format}$, the model must wrap the reasoning process in \token{<think></think>} and the final conclusion in \token{<answer></answer>}. The format reward is 1 if the model complies with this specification and 0 otherwise. For the accuracy reward $\sigma_{acc}$, we extract the answer from \token{<answer></answer>} and compare it with the ground truth. The accuracy reward is 1 if the extracted answer matches the ground truth and 0 otherwise.

\noindent \textbf{Hallucination Reduction Gating.} The correctness of the diagnostic conclusion is paramount. To discourage responses that appear superficially reasonable but yield wrong conclusions, we introduce a hallucination reduction gate. It sets our proposed reward to zero when the final answer is incorrect. The gating mechanism is formulated as $\delta(\cdot) = \mathbf{1}\{\sigma_{acc}=1\}$.

\noindent \textbf{Overall Reward.} We combine all rewards with reweighting to balance the reward contribution using linear weights $\lambda$. The final reward calculation is shown as~\cref{eqn:allrwd}.

{
\vspace{-1.5em}
\begin{equation}
\begin{aligned}
\sigma&=\lambda_{format}\sigma_{format}+\lambda_{acc}\sigma_{acc} \\
& +\delta(\lambda_{PQlR}\sigma_{PQlR}+\lambda_{PQtR}\sigma_{PQtR}+\lambda_{ESR}\sigma_{ESR})
\label{eqn:allrwd}
\end{aligned}
\end{equation}
\vspace{-1em}
}

\subsubsection{Training Procedure}

\noindent \textbf{Reinforcement Learning Algorithm.} We adopt Group Relative Policy Optimization (GRPO)~\cite{shao2024deepseekmath} as the reinforcement learning strategy, employing the proposed reward functions to enhance the model’s exploration capability in complex echocardiography analysis, as illustrated in~\cref{fig:main}.a. The complete GRPO objective used for optimization is calculated as shown in Supplementary~\cref{supp:grpo}.

\noindent \textbf{Two Stage Training Strategy.} To encourage extensive procedure aligned reasoning while preserving diagnostic performance, we adopt a two stage training strategy~\cite{wang2025proactive,rui2025improving}. We base all training on the Qwen 2.5-VL-7B model~\cite{bai2025qwen2}. In stage one, the model is trained with the format, accuracy, and procedural quantity rewards. This stage consolidates core echocardiography recognition across multiple views and modalities and preserves a stepwise reasoning structure. We set $\lambda_{format}=1$, $\lambda_{acc}=1$, $\lambda_{PQtR}=1$ in this stage. In stage two, we apply the complete reward model during training to align the model’s reasoning trajectory with anchor diagnostic procedures. We set $\lambda_{format}=1$, $\lambda_{acc}=1.5$, $\lambda_{PQlR}=0.8$, $\lambda_{PQtR}=0.5$, and $\lambda_{ESR}=0.5$ in this stage. 

\subsection{Template-guided Reasoning Rectification}
\label{sec:inference}
Building on CRT, we propose Template guided Reasoning Rectification (TRR) for scalable inference in \methodname{}. As~\cref{fig:main}.d shows, during inference we first retrieve a reasoning template from CRT using the disease query. The trained model then follows this template to generate stepwise reasoning. \Cref{fig:main}.c shows an example of the reasoning process with the complete echocardiography study and the retrieved template as inputs.

After the model completes reasoning, we reuse the verifier and scoring rule from the Procedural Quality Reward (\cref{sec:reward}) as the reasoning step verifier. This verifier scores each step against the retrieved template and outputs a step quality score. If the average step quality exceeds a preset threshold, we accept the conclusion as compliant with standard procedures and return it as the final output. If the average step quality is below the threshold, we prompt the model again using the previous reasoning path and highlight steps with low template adherence. The prompt instructs the model to reconsider and attend to these flagged elements. We then return the revised conclusion as the final output. We empirically set the threshold to 0.4 in our experiments.

\begin{table*}[htb]
\caption{Performance on echocardiographic diagnosis benchmarks. Name with ``$\star$'' indicates that the model was tuned on the training set. Other models are evaluated directly using the official checkpoints. ``-ICL'' denotes providing the model with retrieved CRT as input at inference only. ``-SFT'' denotes supervised finetuning without CRT. ``-GRPO'' denotes GRPO tuning with accuracy and format rewards~\cite{guo2025deepseek} without CRT. ``RQ'' denotes the Reasoning Quality score. We use GPT-5, following the common evaluation protocol~\cite{wang2025proactive,qiu2025quantifying}, to compute scores in the range $[0,5]$. ``-'' denotes no reasoning process in the inference thus unavailable. ``Ours'' denotes results with both CRT and \methodname{}. Best results are in \textbf{bold}. Second bests are in \underline{underline}. For all metrics, higher is better. }
\label{tab:main}
\tiny
\centering
\renewcommand{\arraystretch}{0.85}
\begin{adjustbox}{width=0.9 \textwidth}
\begin{tabular}{lccccccc}
\toprule
\multicolumn{1}{c}{\multirow{2}{*}{Method}} & \multicolumn{3}{c}{EchoComplex}                                                        & \multicolumn{2}{c}{CardiacNet-ASD} & \multicolumn{2}{c}{CardiacNet-PAH} \\ \cmidrule{2-8} 
\multicolumn{1}{c}{}                        & Accuracy  & F1   & RQ & Accuracy            & F1           & Accuracy            & F1           \\ \midrule
\multicolumn{8}{c}{Echocardiography Foundation Model}                                                                                                                                                          \\ \midrule
EchoPrime~\cite{vukadinovic2025comprehensive}                                   & 0.58                & 0.66     & -                & 0.43                    & \underline{0.57}             & 0.46                    & 0.42            \\
$\text{PanEcho}^\star$~\cite{holste2025complete}                                     & 0.52                & 0.10     & -                                                            & \underline{0.56}                    & 0.09             & 0.31                    & 0.07             \\ \midrule
\multicolumn{8}{c}{General MLLM}                                                                                                                                                                               \\ \midrule
Qwen2.5-VL~\cite{bai2025qwen2}                                  & 0.49        & 0.15 & 3.09       & 0.53                    & 0.44             & 0.31                    & 0.03             \\
Qwen2.5-VL-ICL~\cite{bai2025qwen2}                              & 0.57        & 0.46 & 4.02  & 0.53                    & 0.45             & 0.31                    & 0.05             \\
$\text{Qwen2.5-VL-SFT}^\star$~\cite{bai2025qwen2}                              & 0.76        & 0.69 & -  & 0.42                    & \textbf{0.59}             & 0.3                    & 0.01             \\
$\text{Qwen2.5-VL-GRPO}^\star$~\cite{bai2025qwen2}                             & 0.80         & 0.77 & 3.69         & 0.55                    & 0.27             & 0.31                    & 0.04             \\
$\text{R1-VL}^\star$~\cite{zhang2025r1}                                       & 0.72        & 0.68 & 3.01        & \textbf{0.57}                    & 0.20             & 0.31                    & 0.04             \\ \midrule
\multicolumn{8}{c}{Generalist Medical MLLM}                                                                                                                                                                    \\ \midrule
Lingshu~\cite{xu2025lingshu}                                     & 0.56        & 0.51 & 3.11                                                            & 0.45                    & 0.40             & 0.65                    & 0.77             \\
$\text{Lingshu-GRPO}^\star$~\cite{xu2025lingshu}                                & 0.78       & 0.79 & 3.07                                                            & 0.42                    & \underline{0.57}             & 0.66                    & 0.78             \\
MedGemma~\cite{sellergren2025medgemma}                                    & 0.53       & 0.46 &  3.20                                                           & 0.54                    & \underline{0.57}             & 0.37                    & 0.31             \\
HuatuoGPT-Vision~\cite{chen2024towards}                                & 0.54                & 0.46     &     2.17                                                        & 0.53                    & 0.31             & 0.40                    & 0.32                        \\ \midrule
\multicolumn{8}{c}{Reasoning Medical MLLM}                                                                                                                                                                     \\ \midrule
MedVLM-R1~\cite{pan2025medvlm}                                   & 0.47        & 0.43 &   1.48                                                          & 0.44                    & 0.41             & 0.36                    & 0.42             \\
Chiron-o1~\cite{sun2025enhancing}                                   & 0.53        & 0.44 &     2.76                                                        & 0.40                    & 0.34             & 0.47                    & 0.58             \\
Med-RwR~\cite{wang2025proactive}         &     0.49          &   0.48   &         3.00                                                    & 0.52                    & 0.56             & 0.47                    & 0.50             \\
\textbf{Ours}                                  & \underline{0.83}        & \underline{0.81} &  \underline{4.40}                                                           & 0.48                    & \textbf{0.59}             & \underline{0.67}                    & \textbf{0.79}             \\
\textbf{Ours+TRR}                              & \textbf{0.84}         & \textbf{0.82} &  \textbf{4.48}                       & 0.49                    & \textbf{0.59}             & \textbf{0.68}                    & \underline{0.78}             \\ \bottomrule
\end{tabular}
\end{adjustbox}
\vspace{-2em}
\end{table*}

\section{Experiments}
\label{sec:exp}

\subsection{Experiment Settings}

\noindent \textbf{Development Data and EchoComplex.} To the best of our knowledge, no public echocardiography dataset spans multiple views and modalities for diagnosing complex cardiac diseases\footnote{MIMIC-IV-Echo~\cite{gow2023mimic} has regional ban and incomplete echocardiography study record, therefore unavailable for this study.}. We therefore curated a training dataset and a test dataset from our collaborating hospital. The training dataset contains 1,486 patients covering 15 disease categories. Each patient has a complete echocardiographic study with annotated view labels, quantitative measurements, and a diagnostic report confirmed by cardiologists. Disease labels were derived from these reports. In total, we constructed 2,550 labeled training instances with both positive and negative examples. To benchmark models' ability for multiview, multimodal echocardiographic reasoning at complex diseases, we curate a test set, EchoComplex, comprising 623 patients across the same 15 categories. It matches the data composition of the training dataset. From it, we created 846 diagnostic questions about cardiac diseases for testing. Institutional Review Board approved this disidentified study. Further details are provided in Supplementary~\cref{supp:dataset}.

\noindent \textbf{Public Benchmarks.} We further evaluate generalization to single view echocardiographic diagnosis using two public benchmarks, CardiacNet-ASD and CardiacNet-PAH~\cite{cardiacnet2025}. They provide apical four chamber (A4C) echocardiography videos for diagnosing pulmonary hypertension (PAH) and atrial septal defect (ASD).

\noindent \textbf{Baseline Methods.} We conduct extensive comparisons across four categories of baseline models.
(1) Echocardiography foundation models, which are discriminative models pretrained on large echocardiography datasets.
(2) General multimodal large language models (MLLMs) designed for general purpose visual question answering.
(3) Generalist medical MLLMs trained on diverse multimodal medical data.
(4) Reasoning oriented medical MLLMs optimized for explicit reasoning over medical images.
We provide implementation details for baselines at Supplementary~\cref{supp:baseline}.

\subsection{Main Results}
As shown in~\cref{tab:main}, our model attains the highest accuracy and F1 score on the multiview EchoComplex dataset and the highest F1 score on the single view CardiacNet benchmarks. Compared with echocardiography foundation models, our method yields higher accuracy on EchoComplex.
\methodname{} also surpasses both state-of-the-art medical MLLMs and their reasoning variants on EchoComplex, improving accuracy by 48\% compared with medical MLLMs (0.56 vs. 0.83) and by 6\% compared with trained reasoning variants. 
Automatic reasoning quality assessment~\cite{wang2025proactive,qiu2025quantifying} on the reasoning paths for EchoComplex further indicates that \methodname{} achieves the highest reasoning quality, with a 19\% improvement over all baselines. Details of the assessment are provided in Supplementary~\cref{supp:rq_eval}. 
Notably, even without additional training, directly providing the retrieved template as in-context knowledge at inference increases the base model’s accuracy from 0.49 to 0.57 and its reasoning quality from 3.09 to 4.02, indicating the effectiveness of CRT with canonical stepwise guidance and knowledge. These results highlight that \methodname{} can leverage informative templates from CRT for multi-evidence integrated reasoning similar to that of a cardiologist, which is a key factor for clinical trustworthiness and accuracy. At the same time, medical reasoning MLLMs show reduced reasoning quality on echocardiography analysis, suggesting that multiview multimodal echocardiographic reasoning remains challenging.

\subsection{User Study on Our Results}
We further invited a cardiologist to conduct a preference study. 
We randomly sampled \methodname{} and LingShu~\cite{xu2025lingshu}'s reasoning for 30 patients. 
Blinded to model identity, the cardiologist compared paired outputs and selected the superior reasoning path based on three predefined criteria.
Details of the user study are provided in Supplementary~\cref{supp:user_study}.
The results in~\cref{fig:user_study} show that 93.33\% of \methodname{}’s reasoning trajectories were judged as ``more aligned with Cardiologist Logic''. 73.33\% were preferred for being more ``Clear and Deductive''. 80\% were preferred for stronger ``View and Measurement Involvement''. 
These findings indicate that \methodname{} aligns more closely with the clinical diagnostic workflow and is preferred by clinicians.

{
\begin{figure}[htb]
	\centering
	\includegraphics[width=1.0\columnwidth]{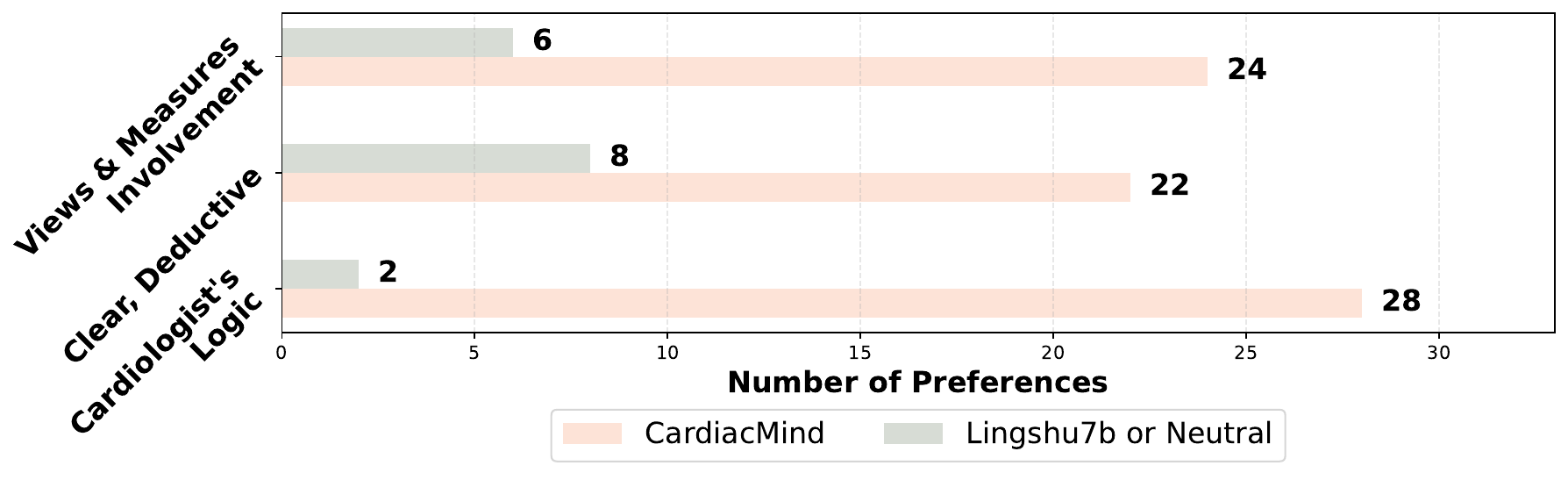}
	\caption{Cardiologist preference result on reasoning quality between our method and LingShu~\cite{xu2025lingshu}. We assess reasoning quality using three criteria. ``Cardiologists Logic'': alignment between the reasoning path and cardiologist diagnostic logic. ``Clear and Deductive'': clarity and deductive structure of the reasoning process. ``View \& Measurement Involvement'': explicit use of views and measurements in the reasoning process. A higher preference number indicates that cardiologists judged the reasoning to be of higher quality.}
	\label{fig:user_study}
    \vspace{-1.5em}
\end{figure}
}

\subsection{Ablative Study}

\begin{table*}[htb]
\caption{Ablation study of reward functions. All configurations that with CRT include accuracy and format rewards with the retrieved template. Ablations with CRT are performed starting from the checkpoint trained in the first stage. Configuration without CRT use accuracy and format rewards to train two epochs. ``RQ'' denotes the Reasoning Quality score consistent with~\cref{tab:main}. Best results are in \textbf{bold}.}
\label{tab:reward_abl}
\tiny
\centering
\renewcommand{\arraystretch}{0.8}
\begin{adjustbox}{width=0.95 \textwidth}
\begin{tabular}{ccccccccccc}
\toprule
\multicolumn{4}{c}{Rewards} & \multicolumn{3}{c}{EchoComplex} & \multicolumn{2}{c}{CardiacNet-ASD} & \multicolumn{2}{c}{CardiacNet-PAH} \\ \midrule
 CRT & PQtR     & PQlR    & ESR    & Accuracy        & F1  & RQ  & Accuracy            & F1           & Accuracy            & F1           \\ \midrule
\no & \no        & \multicolumn{1}{c|}{\no}       & \no      & 0.81                   & 0.78  & 3.48      & \textbf{0.55} & 0.27             & 0.31                    & 0.03          \\
\yes & \no        & \multicolumn{1}{c|}{\no}       & \no      & 0.81                   & 0.79  & 3.78      & \textbf{0.55}  & 0.29             & 0.31                    & 0.05             \\
\yes &\yes        & \multicolumn{1}{c|}{\no}       & \no      & 0.82               & 0.8     & 4.06      & 0.46                    & 0.56             & 0.33                    & 0.21             \\
\yes &\yes        & \multicolumn{1}{c|}{\yes}       & \no      & 0.81                 & 0.81    & 4.26      & 0.46                    & 0.56             & 0.66                    & 0.75             \\
\yes &\yes        & \multicolumn{1}{c|}{\yes}       & \yes      & \textbf{0.83}                & \textbf{0.81}     & \textbf{4.40}      & 0.48                    & \textbf{0.59}             & \textbf{0.67}                    & \textbf{0.79}              \\ \bottomrule
\end{tabular}
\end{adjustbox}
\vspace{-1em}
\end{table*}

\subsubsection{CRT and Reward Analysis}

We first ablate the proposed reward functions to evaluate their effects on reasoning trajectories and accuracy. The results are shown in~\cref{tab:reward_abl}. 
When only accuracy and format rewards are applied, the model maintains decent accuracy on EchoComplex (81\%). However, it produces short reasoning trajectories with limited diagnostic insight with reasoning quality score dropping to 3.78. 
When PQtR and PQlR are applied, both reasoning quality and diagnosis F1 score increase (F1: 0.79 to 0.81, reasoning quality: 3.78 to 4.26). This indicates that PQtR and PQlR guide the model to reason logically and to adhere to standard diagnostic procedures. The best reasoning quality and accuracy are obtained when all three rewards are applied, further demonstrating the effectiveness of the echocardiography semantic reward. Altogether, the three rewards in \methodname{} encourage the model to use this prior knowledge to address critical visual centric challenges and to improve disease diagnosis.
Moreover, models trained with basic rewards without CRT yield worse reasoning quality, showing the importance of the canonical stepwise knowledge introduced by CRT.

{
\begin{table}[htb]
\caption{Ablation for Procedural Quality Reward on EchoComplex. RQ denotes the Reasoning Quality score, consistent with~\cref{tab:main}. ``Similarity-Based'' maximizes the embedding similarity between the first sentence of each reference step and each step in the model's reasoning. For ``Paragraph Verify'', we feed each step's descriptive content to the LLM judge rather than converting it into actionable questions.}
\label{tab:pqlr}
\tiny
\centering
\renewcommand{\arraystretch}{0.9}
\begin{adjustbox}{width=0.85 \columnwidth}
\begin{tabular}{cccc}
\toprule
                    Design    & Accuracy  & F1        & RQ \\ \midrule
Similarity-Based        & 0.82          & 0.81          & 4.31                                                            \\
Paragraph Verify        & 0.81          & 0.80          & 4.23                                                            \\
CardiacMind             & \textbf{0.83} & \textbf{0.81} & \textbf{4.40}                                                   \\ \bottomrule
\end{tabular}
\end{adjustbox}
\vspace{-1em}
\end{table}
}
\subsubsection{Procedural Quality Reward Analysis}
The Procedural Quality Reward (PQlR) is the core reward that incentivizes the model to reason following canonical steps and to aggregate heterogeneous information. 
We therefore evaluate and validate key design choices in PQlR. We consider several alternatives.
The first directly maximizes embedding similarity between the first sentence of the reference template's step and each model step. The similarity is computed based on Qwen3-Embedding-0.6B~\cite{zhang2025qwen3} embeddings.
The second feeds the original descriptive sentences to the verifier instead of using actionable questions. 
\Cref{tab:pqlr} summarizes the comparison, and Supplementary~\cref{supp:abl_demo} provides representative reasoning outputs for these choices.
All alternatives yield comparable diagnosis performance on EchoComplex. This shows that PQlR is a robust way to leverage the high-level content in CRT and to encourage reasoning with multisource information integration. However, using actionable questions with the LLM as Judge scheme yields the best reasoning quality, with a 4\% increase. 
This supports \methodname{}’s design choice of verifying content through direct questions. This strategy simplifies the evaluation of logic-intensive content while still leveraging the high-level semantics provided by the template.

\subsubsection{Training Procedure Analysis}

We further assessed the necessity of two training components: Hallucination Reduction Gating (HG) and the Verbose Tolerance Threshold (VT). \Cref{tab:traintrick} shows that removing HG reduces accuracy to 0.79. This suggests that when the model arrives at an incorrect conclusion, the three additional rewards risk diluting the accuracy signal, which HG is designed to prevent. Training without VT causes a slight performance drop. This likely occurs because the model outputs more reasoning steps to pursue a higher PQlR score but fails to effectively aggregate information from verbose outputs. These results validate the necessity of both components.
\vspace{-0.5em}
{

\begin{table}[htb]

\caption{Ablation study of \methodname{} training procedures on EchoComplex. VT denotes the Verbose Tolerance Threshold, and HG denotes Hallucination Reduction Gating.}
\label{tab:traintrick}
\tiny
\centering
\renewcommand{\arraystretch}{0.75}
\begin{adjustbox}{width=0.7 \columnwidth}
\begin{tabular}{ccccc}
\toprule
VT & HG & Accuracy        & F1    \\ \midrule
\yes                                                           & \no                                                              & 0.79                     & 0.77      \\
\no                                                           & \yes                                                              & 0.81                      & 0.79      \\
\yes                                                           & \yes                                                             & \textbf{0.83}                     & \textbf{0.81}      \\ \bottomrule
\end{tabular}
\end{adjustbox}
\vspace{-1em}
\end{table}
}
\vspace{-.5em}
\subsubsection{Template-guided Reasoning Rectification Analysis}
We also analyzed the performance gains attributable to Template guided Reasoning Rectification. As shown in~\cref{tab:main}, both accuracy and F1 score increase after rectifying the reasoning trajectories for cases with initially low reasoning quality scores. Comparative case studies of the rectification process are provided in Supplementary~\cref{supp:case_study}. With the reconsideration prompt, the model explicitly identifies and assesses critical steps that the previous round of reasoning may have neglected. It then revises the final conclusion, which leads to better results.

\section{Conclusion}
\label{sec:conc}
\vspace{-.5em}
In this paper, we present the Cardiac Reasoning Template (CRT) and \methodname{}, a novel approach that systematically embeds cardiologist-like mindset into medical reasoning MLLMs. We curate a library of concise stepwise echocardiographic diagnostic procedures for 15 complex cardiac diseases from authoritative sources. We design a two stage rule based reinforcement framework with three new rewards that promote comprehensive analysis of echocardiography studies and faithful adherence to CRT. We also use CRT to scale inference by assessing and correcting the reasoning process. Experiments on 15 echocardiographic diagnostic tasks, external validation, and clinical ratings demonstrate the effectiveness of our method.

{
    \small
    \bibliographystyle{ieeenat_fullname}
    \bibliography{bib.bib}
}

\clearpage
\setcounter{page}{1}
\maketitlesupplementary
\appendix

\section{Data Construction Details}
\label{supp:dataset}
To curate a multiview, multimodal echocardiography dataset for complex cardiac disease diagnosis, we first retrieved patient records from 2022 to 2025 that included 15 targeted types of complex cardiac diseases as well as patients without detectable abnormalities. \Cref{tab:diseases} lists all complex cardiac disease categories included in our study. In total, we retrieved 2109 patients.
Each patient record comprises a complete echocardiography record, quantitative cardiac measurement data, and a clinical echocardiography report. The complete echocardiography record contains transthoracic echocardiography (TTE) 2D imaging, Color Doppler, M-Mode, Continuous Wave (CW) Doppler, and Pulse Wave (PW) Doppler acquisitions from multiple standard cardiac views. On average, each patient has 54 echocardiography records. This comprehensive study design mirrors the information available to cardiologists in routine clinical practice and is sufficient to support reliable echocardiography based diagnosis.
We ensured that there is no overlap between the patients in the training and testing sets.
We followed prior work~\cite{holste2025complete,vukadinovic2025comprehensive} to further process the raw echocardiography videos. All echocardiography videos are standardized to a spatial resolution of $224\times224$ with 16 frames. All static contents (e.g., M-Mode and CW/PW Doppler) are resized such that the shortest side is $224$ while preserving the original aspect ratio.

\begin{table*}[htb]
\caption{Selected diseases and their corresponding sample counts in the training set and EchoComplex. SVC: Superior Vena Cava. ASD: Atrial Septal Defect. VSD: Ventricular Septal Defect. PAH: Pulmonary Hypertension. HFpEF: Heart Failure with preserved Ejection Fraction. Normal indicates that no abnormality is identified in the patient's echocardiographic report.}
\label{tab:diseases}
\tiny
\centering
\renewcommand{\arraystretch}{0.85}
\begin{adjustbox}{width=0.9 \textwidth}
\begin{tabular}{cccccc}
\toprule
Disease Name                & Training Set & EchoComplex             & Disease Name         & Training Set & EchoComplex \\ \midrule
Ischemic Cardiomyopathy     & 118          & \multicolumn{1}{c|}{54} & Outlet VSD           & 38           & 10          \\
Hypertrophic Cardiomyopathy & 102          & \multicolumn{1}{c|}{59} & Muscular VSD         & 23           & 5           \\
Dilated Cardiomyopathy      & 116          & \multicolumn{1}{c|}{57} & Perimembranous VSD   & 54           & 10          \\
Rheumatic Cardiomyopathy    & 20           & \multicolumn{1}{c|}{10} & Atrial Myxoma        & 35           & 5           \\
Ostium Secundum ASD         & 112          & \multicolumn{1}{c|}{50} & Left Atrial Thrombus & 34           & 5           \\
SVC type Sinus Venosus ASD  & 15           & \multicolumn{1}{c|}{10} & Ventricular Aneurysm & 62           & 20          \\
PAH                         & 192          & \multicolumn{1}{c|}{89} & Fabry                & 72           & 15          \\
HFpEF                       & 113          & \multicolumn{1}{c|}{50} & Normal               & 380          & 174         \\ \bottomrule
\end{tabular}
\end{adjustbox}
\end{table*}

\section{Cardiac Reasoning Template Construction}
\label{supp:CRT}

\begin{figure}[htb]
	\centering
	\includegraphics[width=1.0\columnwidth]{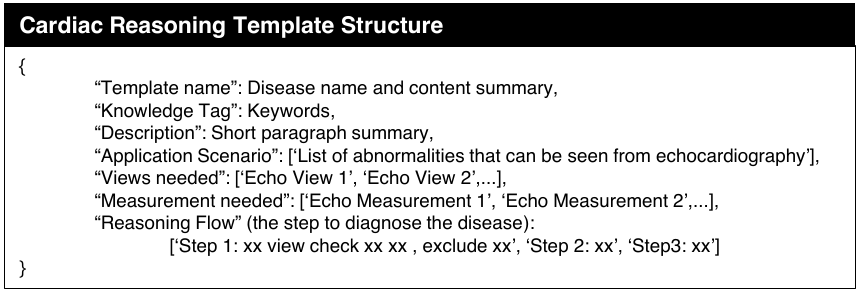}
	\caption{The template structure of Cardiac Reasoning Template.}
	\label{fig:crt_structure}
\end{figure}

\noindent \textbf{Cardiac Reasoning Template Construction Details}
We curated the Cardiac Reasoning Template (CRT) from authoritative echocardiography knowledge sources, including a standard textbook~\cite{oh2006echo} and disease specific clinical guidelines for each selected condition~\cite{lang2006recommendations,mitchell2019guidelines,kittleson20232023,sirnes2009guidelines,minette2006ventricular,arbelo20232023,ommen20242024,heidenreich20222022,stout20192018,silvestry2015guidelines,wiegers20192019}. The textbook provides fundamental echocardiography principles, whereas the guidelines describe up to date diagnostic techniques for the corresponding diseases. We used GPT-5 to organize the CRT entries into structured templates, as illustrated in~\cref{fig:crt_structure}.
The constructed CRT contains 42 templates in total. Each template has on average 7 steps, which reflects the necessity of multi step reasoning for complete and rigorous diagnosis. On average, each template consists of 335 words and specifies detailed echocardiographic views and measurements, thereby encoding rich cardiology knowledge.
After constructing the CRT, we employed Qwen3-Embedding~\cite{zhang2025qwen3} to retrieve templates by selecting the one with the highest embedding similarity between the disease query and the template metadata.

\begin{figure*}[htb]
	\centering
	\includegraphics[width=1\textwidth]{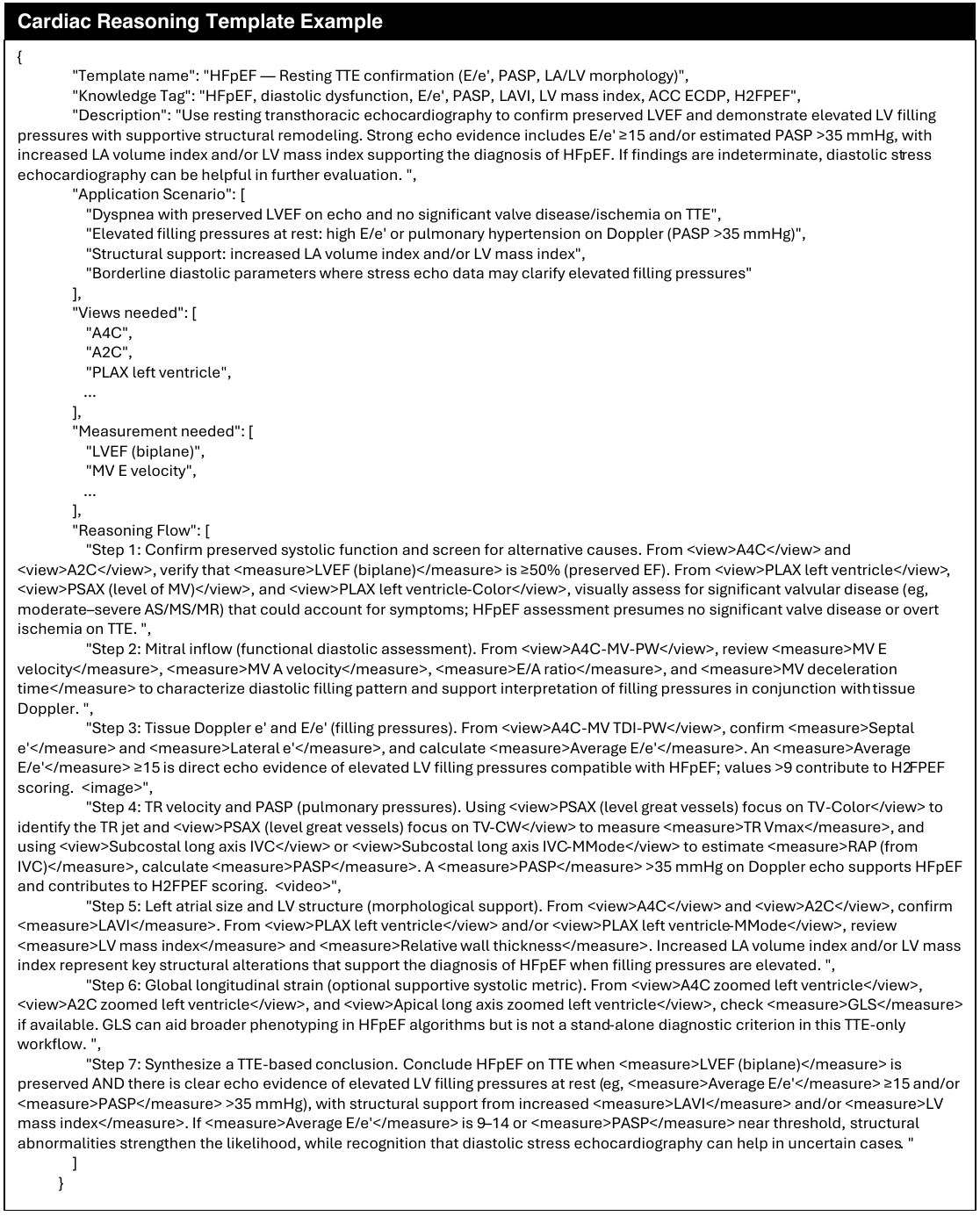}
	\caption{Example of a template in the Cardiac Reasoning Template suite. In ``Views Needed'' and ``Measurement Needed'' sections, we shorten the lists of views and measurements for brevity, since the corresponding view and measurement names are already sufficiently specified in the ``Reasoning Flow'' section.}
	\label{fig:crt_example}
\end{figure*}

\noindent \textbf{Examples of Cardiac Reasoning Template.} \Cref{fig:crt_example} presents a complete example template from the CRT. The template provides detailed step by step instructions for the reasoning MLLM to execute diagnostic procedures, such as confirming abnormalities, comparing with normal reference values, and formulating intermediate findings. In this way, the MLLM is explicitly guided to follow the intended diagnostic workflow, while \methodname{}'s reward design incentivizes the MLLM to reason toward the correct conclusion based on these well defined steps.

\section{Procedural Quality Reward Implementation Detail}

\subsection{Question List Conversion Implementation}
\label{supp:qlist}
\begin{figure*}[htb]
	\centering
	\includegraphics[width=0.9\textwidth]{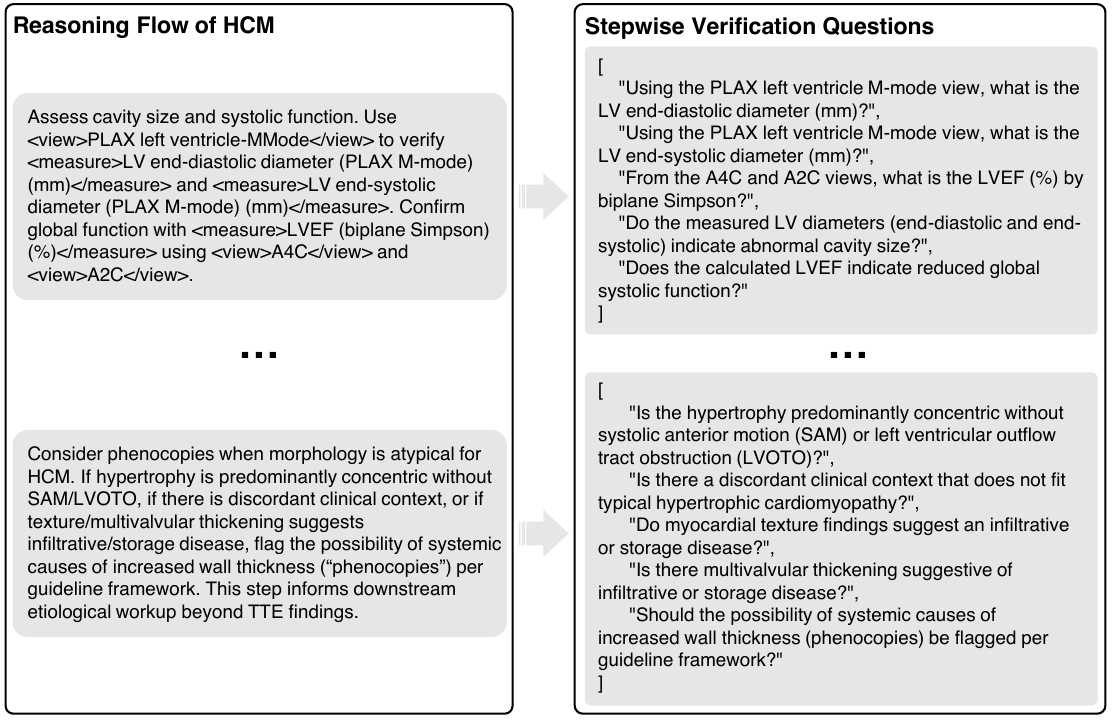}
	\caption{Example of converting steps in the reasoning flow of the Cardiac Reasoning Template into question lists. HCM: Hypertrophic Cardiomyopathy.}
	\label{fig:pqlr_question_example}
\end{figure*}

Procedural Quality Reward must verify whether \methodname{}'s reasoning steps correctly address the key diagnostic procedures specified in the reasoning template, which imposes stringent requirements on the verifier to understand the logic and context in both the template and the model response. 
To simplify the evaluation of such logic intensive content, while still leveraging the high level diagnostic information encoded in the template, we convert the stepwise verification target from the original descriptive text into a list of questions associated with the analysis of echocardiographic views and measurements. Specifically, for each step in the template, we use GPT-5 to perform this conversion. \Cref{fig:pqlr_question_example} illustrates the conversion process and shows the resulting question list used for verification. After the conversion, for each individual patient, we select the questions whose required views are present in that patient's study, based on the annotated view labels and the view requirements in the template metadata.

\subsection{Verifier Implementation Detail}
\label{supp:pqlr_pmt}
We used Qwen3-8B as the backbone verifier for PQlR and ran it in non thinking mode for efficiency. We instructed the verifier to assess each answer along four dimensions: view and measurement involvement, conclusion involvement, factual accuracy, and answer based relevance, and to output scores only in the range from 0 to 1. The complete prompt used for verification is provided in~\cref{tab:prompt-verifier}.

\begin{table}[htb]
\caption{Prompt for the LLM-as-Judge Verifier used in Procedural Quality Reward.}
\label{tab:prompt-verifier}
\centering
\begin{tabular}{p{0.9\linewidth}}
\toprule
\textbf{Prompt for Procedural Quality Reward} \\
\midrule
\small{You are given a list of authoritative questions and a response generated by an AI model. You are also given a echocardiographic view list available for the AI model to analysis. Your task is to evaluate the response based on its accuracy and answer-based relevance to the questions asked. Provide a score between 0 and 1, where 1 indicates a perfect response that based all the answer on the echocardiographic available, provided affirmative or negative conclusions and addressed the questions, and 0 indicates a answer that is: all based on the unavailable echocardiographic views (indicating hallucination), failed to specifically answer any of the specific questions posed, or vague, or completely irrelevant, or no affirmative or negative conclusion is provided. Return your score only. /no\_think}
\\
\bottomrule
\end{tabular}
\end{table}

\section{Model Implementation Details}

\subsection{Prompt Design}
To guide the model to output an ordered step by step diagnostic reasoning path, we first design a prompt template, as shown in~\cref{tab:sys_prompt}. The model is instructed to perform stepwise reasoning that adheres to the input templates and explicitly incorporates the required views and measurements.

\begin{table}[htb]
\caption{Prompt for \methodname{}.}
\label{tab:sys_prompt}
\centering
\begin{tabular}{p{0.9\linewidth}}
\toprule
\textbf{System Prompt for \methodname{}.} \\
\midrule
\small{You are an expert cardiologist. You are given a series of echocardiographic inputs and a disease query. Your task is to diagnose the disease. You are also given a guideline that help you diagnose the disease. In your reasoning steps, you should think carefully to address the demand and questions raised in each step of the guideline using the input echocardiographic content. DO NOT COPY THE GUIDELINE. You should diagnose step-by-step based on the given echocardiographic inputs. You should output your reasoning process in steps carefully and the final diagnosis. In each step of your reasoning output, include the video name (view name) and the measurement name from the input that you used as evidence, and use the word in the input. If your reasoning conclusion is supportive based on TTE, give ``Yes'' final answer. Base your decision only on current TTE input. The reasoning process and answer are enclosed within \token{<think> </think>} and \token{<answer> </answer>} tags, respectively, i.e., \token{<think>} Step 1: xx; Step 2:xx \token{</think><answer>} answer here (Yes/No) \token{</answer>}}
\\
\bottomrule
\end{tabular}
\end{table}

\subsection{Training Procedure}

\noindent \textbf{Reinforcement Training Target}
\label{supp:grpo}
We used Group Relative Policy Optimization (GRPO)~\cite{shao2024deepseekmath} as the reinforcement learning strategy. 
Specifically, for each training data, we sample a group of rollouts $\{r_i\}_{i=1}^G$ from the policy model $\pi_{\theta_\text{old}}$ and acquire the reward $\sigma_i$. The advantage $\hat{A}_{i,t}$ is estimated as the standard score of the group rewards. The optimization target used for training is given as~\cref{eq:grpo_objective}:
\begin{multline}
\mathcal{J}_{\text{GRPO}}(\theta)
= \mathbb{E}_{x,\, \{r_i\}_{i=1}^G \sim \pi_{\theta_{\text{old}}}(\cdot \mid x, \mathrm{T})} \Bigg\{
\frac{1}{G\,|r_i|} \sum_{i=1}^{G} \sum_{t=1}^{|r_i|}
\\
\min\Big[
\rho_{i,t}\,\hat{A}_{i,t},\;
\operatorname{clip}(\rho_{i,t}, 1-\epsilon, 1+\epsilon)\,\hat{A}_{i,t}
\Big]
\\
- \beta\,
\mathbb{D}_{\mathrm{KL}}\!\Big[
\pi_\theta(\cdot \mid x, \mathrm{T})
\;\big\|\;
\pi_{\theta_{\mathrm{ref}}}(\cdot \mid x, \mathrm{T})
\Big]
\Bigg\}.
\label{eq:grpo_objective}
\end{multline}

Here, $\beta$ is the KL divergence penalty coefficient that controls the degree of exploration by the policy model.

\noindent \textbf{Implementation Settings.} All experiments are conducted on 32 H800 GPUs using SWIFT as the training framework. For GRPO training, we perform full parameter fine tuning with a learning rate of $2e^{-6}$ and $\beta$ set to $5e^{-3}$ in the first training stage and $1e^{-2}$ in the second stage, and optimize the model using DeepSpeed with ZeRO-2.

\section{Evaluation Implementation Details}

\subsection{Baseline Model Implementation}
\label{supp:baseline}

We provide implementation details for all baseline methods in~\cref{tab:main}. All MLLM baselines, except ``Qwen2.5-VL-ICL'', are evaluated without inserting retrieved templates from CRT, in order to benchmark their original echocardiography analysis capability. We use the recommended generation configurations with thinking mode, following the official implementations for each baseline. Below we provide additional explanations for specific baselines for clarity.

\noindent \textbf{EchoPrime~\cite{vukadinovic2025comprehensive}.} EchoPrime is a CLIP~\cite{radford2021learning} based vision language model that predicts diseases through video text similarity matching. Since EchoPrime is trained using large scale single video text contrastive learning, we perform direct video text similarity matching at test time to preserve its visual language representations. We follow the original implementation and recommended practice to perform prediction on all test data. Specifically, we use the original repository to select the diagnostic phrase with the highest embedding cosine similarity as the disease prediction. We perform per disease binary classification to report performance. For example, we choose between the sentences ``Findings consistent with pulmonary hypertension.'' and ``PA systolic pressure is normal.'' and take the one with the higher video text similarity as the final prediction.

\noindent \textbf{PanEcho~\cite{holste2025complete}.} PanEcho is a transformer based multi task predictor for 23 heart diseases that relies on disease specific classification heads for diagnosis. However, the original implementation does not include classification heads for the complex cardiac diseases considered in this paper. Therefore, we use PanEcho's pretrained weights and released fine tuning code and configurations as the backbone, and fine tune 15 new single layer binary classification heads. We then use the fine tuned model to perform prediction on the entire test set.

\noindent \textbf{R1-VL~\cite{zhang2025r1}.} We follow the design of R1-VL~\cite{zhang2025r1} and use a keyword matching approach as the ``StepRAR'' reward function, where the keywords include view names and cardiac measurement terms. All other rewards in R1-VL remain unchanged. We apply this setting to fine tune the Qwen2.5-VL-7B model for a fair comparison.

\noindent \textbf{MedVLM-R1~\cite{pan2025medvlm}.} MedVLM-R1 deploys basic rewards in GRPO to perform reinforcement learning, which overlaps with the setting used in the ``Qwen2.5-VL-GRPO'' row. We therefore perform direct inference without additional fine tuning for brevity.

\noindent \textbf{Chiron-o1~\cite{pan2025medvlm}.} Chiron-o1 proposes an LLM based data construction method for reasoning path generation, followed by supervised fine tuning on the constructed data. However, Chiron-o1 assumes that the MLLM used for reasoning path construction already has strong reasoning ability on the target modality and has been exposed to rich data. This assumption does not hold in the echocardiography domain, where open source MLLMs still need to be explicitly incentivized for accurate echocardiography analysis~\cite{thapa2025well}. Consequently, it is difficult to apply Chiron-o1's data construction method to our data. We therefore directly infer with the official Chiron-o1 checkpoint for testing.

\noindent \textbf{MedRwR~\cite{wang2025proactive}.} MedRwR proposes a method to retrieve external information during reasoning. However, the training rewards in MedRwR require constructing a broad echocardiography video based knowledge database with multi video input, and there is currently no sufficient publicly available dataset to satisfy this requirement. Therefore, fully reproducing the MedRwR training setup falls outside the scope of this paper, and we directly use the released MedRwR implementation for testing.

\subsection{Reasoning Score}
\label{supp:rq_eval}

We apply an LLM-as-Judge strategy to evaluate the quality of the reasoning path. Specifically, we use GPT-5 to assess reasoning paths along five dimensions:
\begin{itemize}
    \item \textbf{Factual Quality.} Are the statements and interpretations medically and factually correct?
    \item \textbf{Coherence and Structural Logic.} Does the reasoning follow a clear, deductive, stepwise logic to confirm the presence of disease, assess chamber and valve morphology, quantify findings, and rule out artifacts and alternative structures?
    \item \textbf{Relevance and Focus.} Does the reasoning stay on topic and directly support the given question and final answer?
    \item \textbf{Informative and Analytic Value.} To what extent does the reasoning process use evidence from the input to analyze and support intermediate conclusions, rather than directly stating these conclusions without justification?
    \item \textbf{Echo Rigor and Validation.} Does the reasoning process explicitly mention echocardiography views and measurements, base its analysis on them, and address uncertainty with appropriate caution, instead of relying on generic statements (for example, describing only “the image” or “the video” without echo specific details)?
\end{itemize}
The final overall score is given on a scale from zero to five. We ran the assessment on all reasoning paths generated for EchoComplex.

\subsection{Reasoning Path Clinical Validation}
\label{supp:user_study}
We further invited a cardiologist to conduct a double-blinded paired-preference user study, following a setting similar to that in~\cite{tanno2025collaboration}, to assess reasoning quality from a cardiology perspective. We first randomly sampled 30 patients from 7 cardiac disease categories in EchoComplex, namely Dilated Cardiomyopathy, HFpEF, Hypertrophic Cardiomyopathy, Ischemic Cardiomyopathy, Ostium Secundum ASD, Rheumatic Cardiomyopathy, and Ventricular Aneurysm. For each selected patient, we used \methodname{} and LingShu-7B to generate both the reasoning path and the final diagnostic conclusion. We then anonymized the model identities and randomly permuted the order of the paired responses so that the cardiologist was blinded to the source. The cardiologist was instructed to select the reasoning path with higher quality according to three criteria. \textbf{Cardiologist’s Logic} evaluates which option is more aligned with standard cardiology diagnostic reasoning. \textbf{Clear and Deductive} evaluates which option presents clearer, more deductive, stepwise logic that confirms the presence of relevant findings, assesses adherence and morphology, quantifies observations, and rules out artifacts or other structures. \textbf{View and Measurement Involvement} evaluates which option more precisely specifies echocardiographic views and measurements, grounds its analysis in these elements, and addresses uncertainty with appropriate caution, rather than relying on generic references (e.g., “the image” or “the video”). For each pair, the cardiologist could choose option A, option B, or a neutral preference.

\section{Template-guided Reasoning Rectification Implementation Detail}

We further performed template-guided reasoning rectification to identify and correct reasoning responses with low procedural adherence to the template, thereby improving diagnostic accuracy. Low template procedural adherence suggests that the reasoning path may deviate from the anchor diagnostic procedure and is therefore more likely to overlook or misinterpret critical information.

To identify low-adherence responses, we first generated a reasoning path for each test case. We then reused the verifier and scoring criteria described in Procedural Quality Rewards to rate the quality of each individual step in every reasoning path. After obtaining the stepwise reasoning quality scores, we averaged them within each case to obtain a single reasoning quality score per test example. Cases whose average score fell below a predefined threshold were subjected to a second round of reasoning.

For these low-scoring cases, we first identified the low-quality reasoning steps by selecting those steps with scores lower than the case-wise average minus one median absolute deviation. We then prompted the model with the previous conclusion and the identified low-quality steps, asking it to reconsider its reasoning and update the conclusion if necessary. The revised conclusion was taken as the final output for that case.

\section{Case Studies}
\label{supp:case_study}
We provide detailed reasoning outputs of \methodname{} with case studies in this section. \Cref{supp:cmd_demo} presents examples of \methodname{}'s reasoning on both diseased and normal patients. \Cref{supp:trr_demo} presents a reasoning example in which \methodname{} is augmented with Template-guided Reasoning Rectification. \Cref{supp:abl_demo} presents reasoning examples used in the ablation studies. Note that all multimodal inputs comprise a complete echocardiography record, and we only illustrate four echocardiographic views in the figures for brevity.

\subsection{Examples of \methodname{}'s Reasoning Path}
\label{supp:cmd_demo}
\begin{figure*}[htb]
	\centering
	\includegraphics[width=1\textwidth]{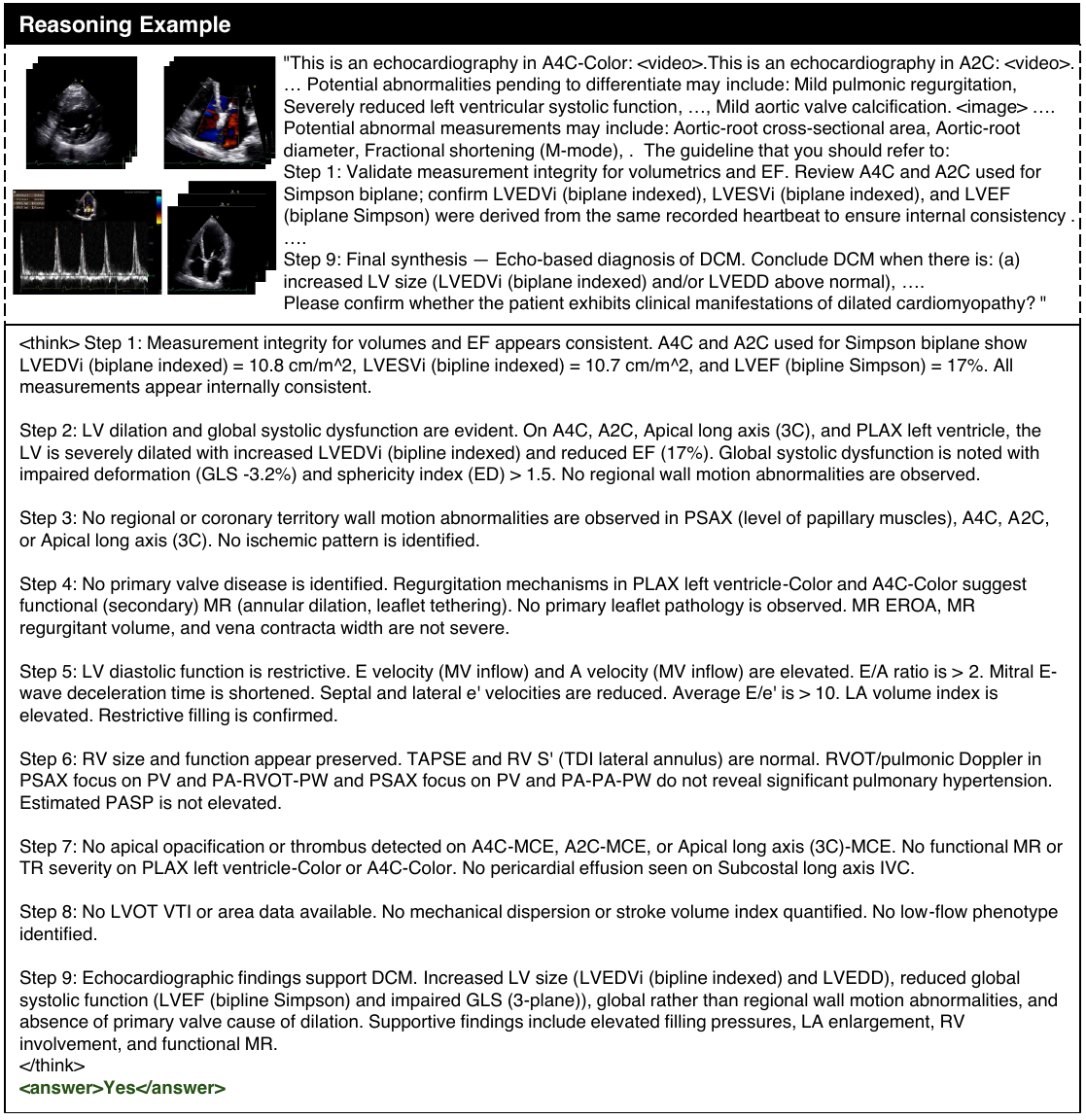}
	\caption{Example of \methodname{}'s reasoning process for a diseased patient with dilated cardiomyopathy. Text in bold green indicates correct statements.}
	\label{fig:casestudy1}
\end{figure*}

\Cref{fig:casestudy1} demonstrates \methodname{}'s reasoning process for a patient with dilated cardiomyopathy. The reasoning explicitly confirms quantitative echocardiographic measurements within the reasoning steps, such as ``LVEDVi (biplane indexed) = 10.8 cm/m2, LVESVi (bipline indexed) = 10.7 cm/m2'', and associates cardiac clinical manifestations with the corresponding measurement abnormalities, for example ``Global systolic dysfunction is noted with impaired deformation (GLS -3.2\%) and sphericity index (ED) $>$ 1.5''. It also specifies the echocardiographic views to inform cardiologists of the anatomical source of each piece of information. In steps 3 and 7, the reasoning process follows guideline-based practice to rule out potential confounders and etiologies (``No ischemic pattern is identified'' and ``No pericardial effusion''). Finally, the reasoning synthesizes these findings into a coherent diagnostic summary. Overall, for diseased patients, the reasoning identifies key abnormalities and explicitly links them to guideline-consistent diagnostic steps that support the final diagnosis.

\begin{figure*}[htb]
	\centering
	\includegraphics[width=1\textwidth]{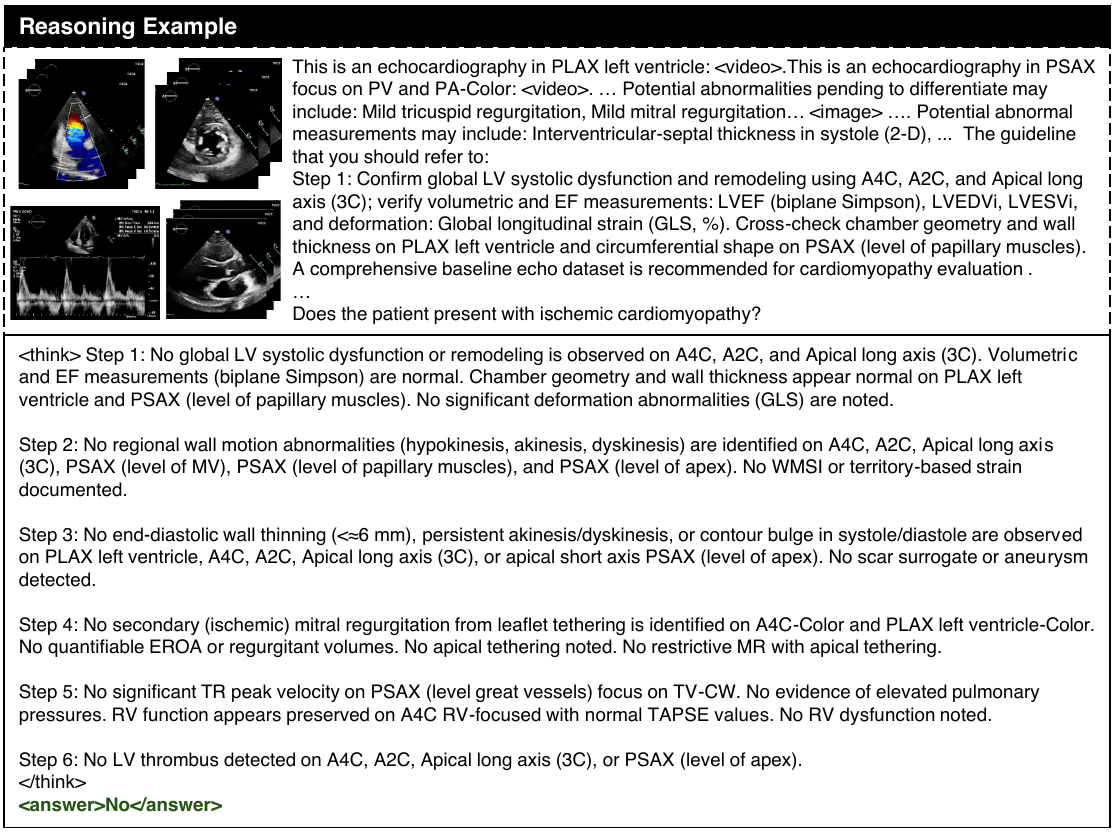}
	\caption{Example of \methodname{}'s reasoning process for a patient without ischemic cardiomyopathy. Text in bold green indicates correct statements.}
	\label{fig:casestudy2}
\end{figure*}

\Cref{fig:casestudy2} shows \methodname{}'s reasoning for a negative patient. Following the reasoning template, the model systematically confirms the absence of abnormalities in a logically structured manner and ultimately rules out the possibility of the queried disease.

\subsection{Examples of \methodname{}'s Reasoning Path with Template-guided Reasoning Rectification.}
\label{supp:trr_demo}

\begin{figure*}[htb]
	\centering
	\includegraphics[width=.9\textwidth]{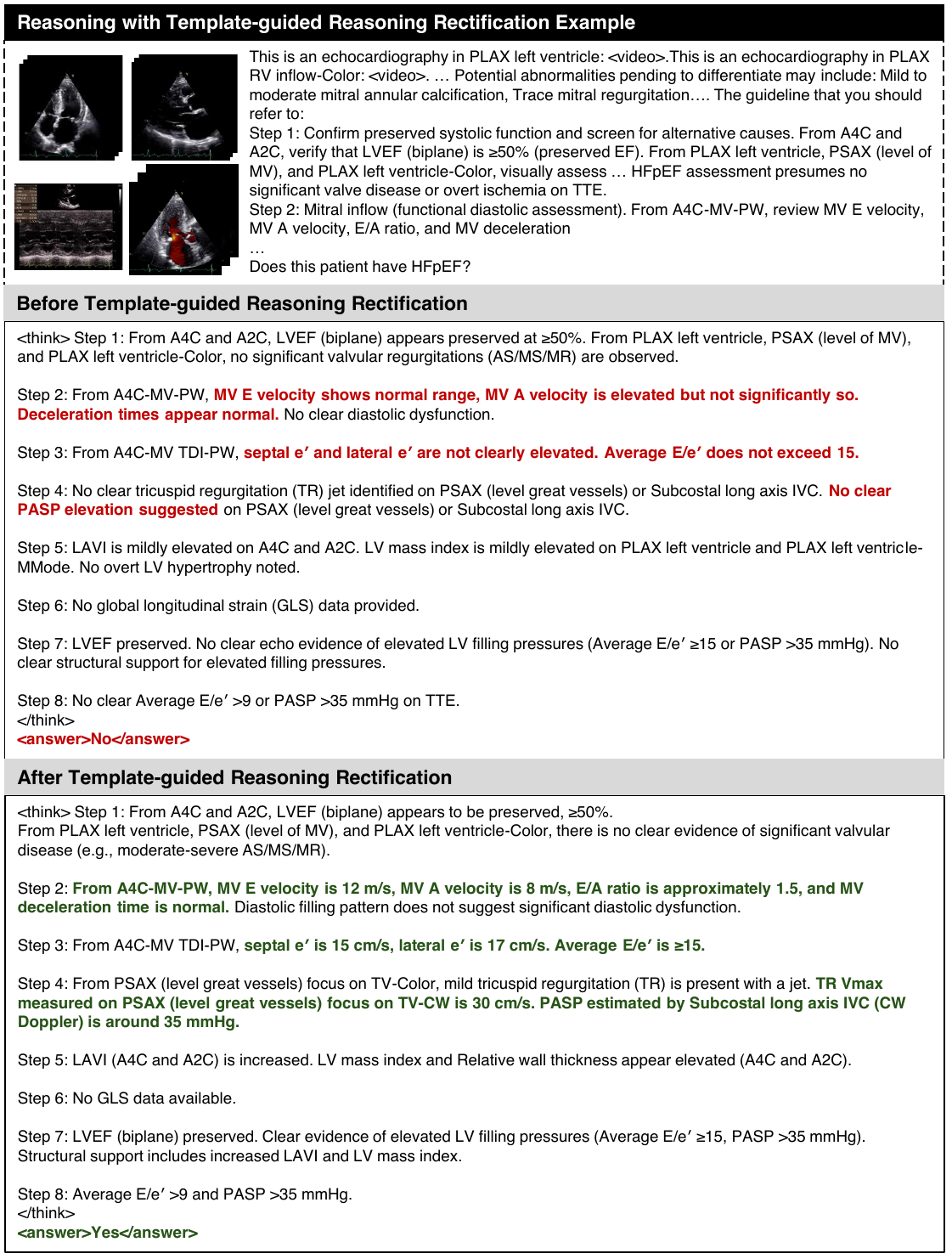}
	\caption{Example of \methodname{}'s reasoning process when augmented with template-guided reasoning rectification. Text in bold green indicates correct statements. Text in bold red indicates ungrounded and vague statements.}
	\label{fig:trrexp}
\end{figure*}

\Cref{fig:trrexp} illustrates the reasoning quality improvement achieved by template-guided reasoning rectification. In the first round of reasoning, the model's reasoning path contains ungrounded and vague statements that lead to incorrect intermediate conclusions, such as ``shows normal range'' and ``appear normal'' in step 2, without referencing any quantitative measurements. This, in turn, results in an incorrect diagnosis. After template-guided reasoning rectification, the model is explicitly informed of the previously low-quality reasoning steps and subsequently provides detailed quantitative measurements before drawing conclusions, as in steps 2, 3, and 4. The improved reasoning process ultimately leads to a revised conclusion and a correct diagnosis.

\subsection{Ablative Case Studies}
\label{supp:abl_demo}

\begin{figure*}[htb]
	\centering
	\includegraphics[width=0.75\textwidth]{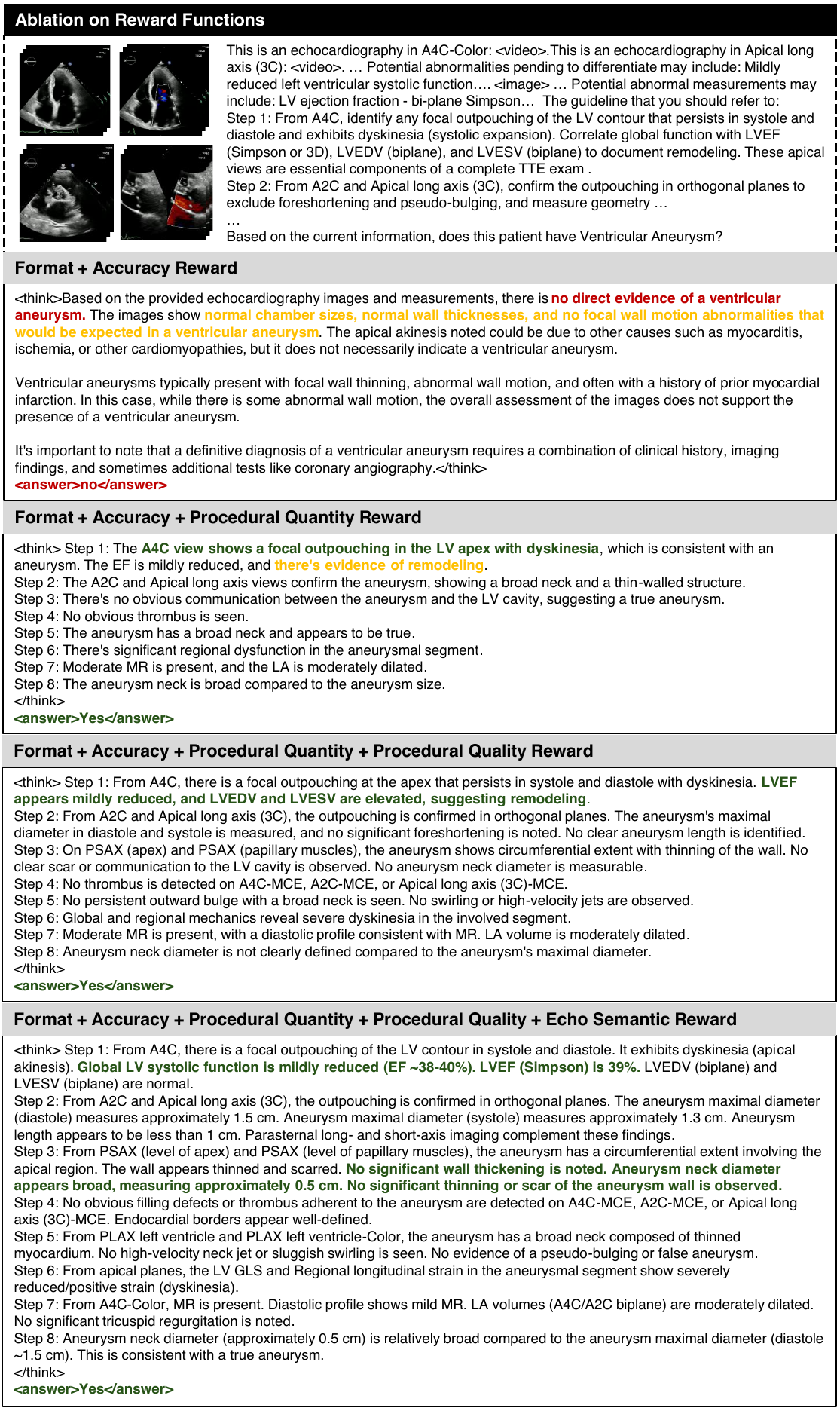}
	\caption{Ablation examples for reward functions. Text in bold green indicates correct statements. Text in bold red indicates incorrect statements. Text in bold yellow indicates brief statements that lack detail. Zoom in for best viewing.}
	\label{fig:rwd_exp}
\end{figure*}

\noindent \textbf{Ablative Examples for Reward Functions.} \Cref{fig:rwd_exp} shows the reasoning results when the reward components are gradually incorporated into the training process. When only the basic accuracy and format rewards are applied, the model produces condensed outputs without explicit logical reasoning or detailed specification of views and measurements. When the procedural quantity reward is added, the reasoning path becomes more structured and follows a clearer logical order. However, the content remains abbreviated and often lacks grounded information, such as the vague phrase ``evidence of remodeling''. When the Procedural Quality Reward is applied, the stepwise reasoning becomes more detailed and includes explicit references to echocardiographic views. When all rewards are applied together, they act synergistically to encourage the model to provide detailed measurements and view annotations, accompanied by observations that support the analysis and final conclusion, as illustrated in steps 1–3. This example highlights the distinct role and advantages of each proposed reward component.

\begin{figure*}[htb]
	\centering
	\includegraphics[width=0.78\textwidth]{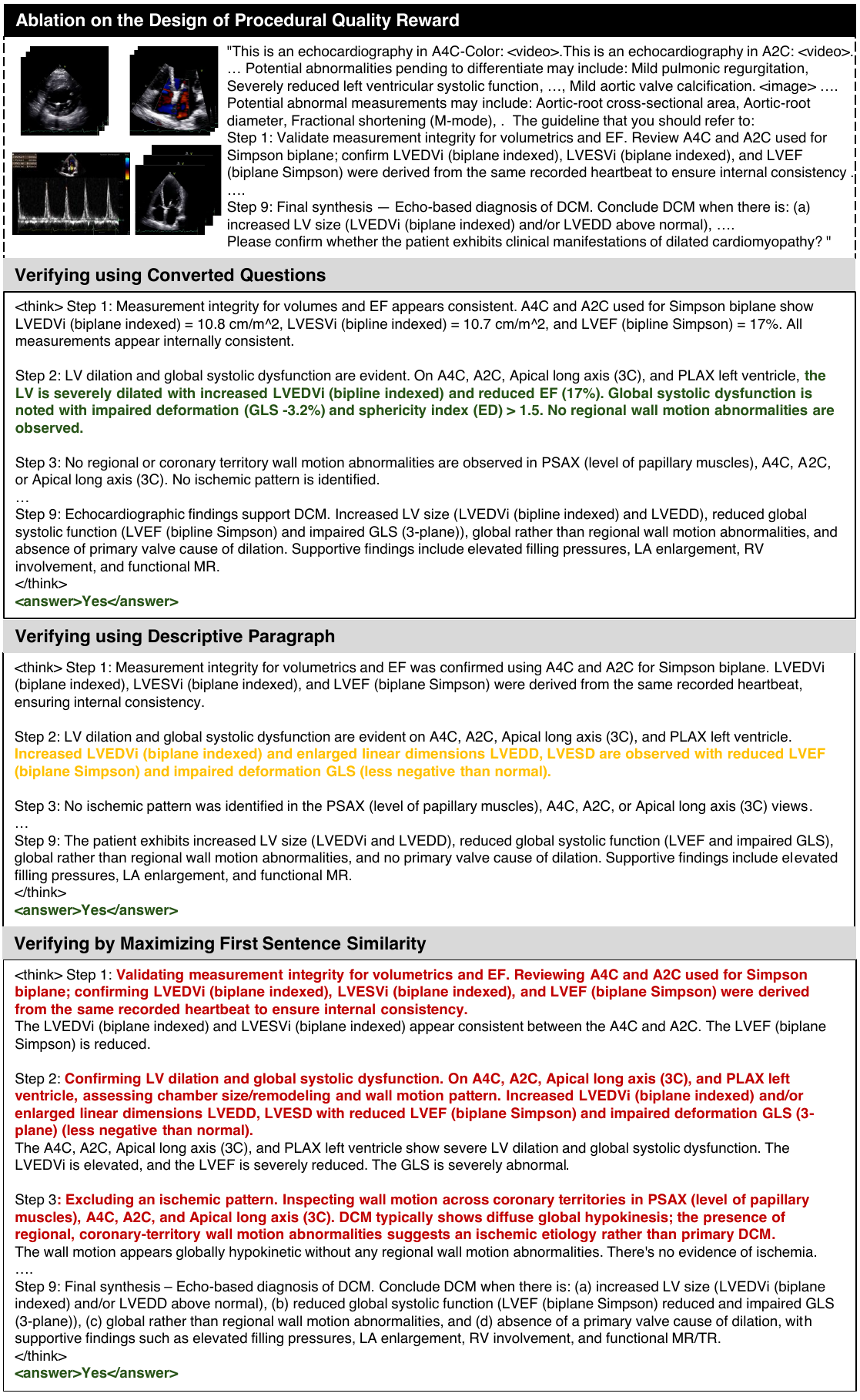}
	\caption{Ablation examples for design choices of the Procedural Quality Reward. Text in bold green indicates correct statements. Text in bold red indicates statements repeated from the given template. Text in bold yellow indicates brief statements that lack detail. Zoom in for best viewing.}
	\label{fig:pqlr_exp}
\end{figure*}

\noindent \textbf{Ablative Examples for the Design Choice of PQlR.} \Cref{fig:pqlr_exp} shows the reasoning results for different design choices of Procedural Quality Rewards. When the model is verified directly against an overall descriptive paragraph, the reward model cannot provide comprehensive feedback on how the reasoning output addresses the critical questions in the given templates, which leads to an abbreviated reasoning process with descriptions not directly addressing vital diagnostic questions. When PQlR uses a simple strategy that maximizes first-sentence similarity to promote contextual relevance, the reasoning model tends to replicate or rephrase the original template (see bold red text in~\cref{fig:pqlr_exp}) without further analysis or additional conclusions. This increases the verbosity of the model while reducing the informativeness of the reasoning process.

\end{document}